\documentclass[runningheads]{llncs}
\usepackage{graphicx}
\usepackage{svg}
\usepackage{pgf}
\usepackage{pgfplots}
\usepackage{lmodern}
\usepackage{listings}
\usepackage{subcaption}
\usepackage{booktabs}
\usepackage{sidecap}
\usepackage{graphicx}
\usepackage{pifont}
\usepackage{amsmath}
\usepackage{amssymb}
\usepackage{array}
\usepackage{placeins}
\usepackage{multirow}
\usepackage{cite}
\usepackage[ruled,vlined,linesnumbered]{algorithm2e}
\usepackage{hyperref} % import the package

\begin{document}
\title{Beyond explaining: XAI-based Adaptive Learning with SHAP Clustering for Energy Consumption Prediction}
\titlerunning{XAI-based Adaptive Learning with SHAP Clustering}
\author{Tobias Clement\inst{1} \and
Hung Truong Thanh Nguyen\inst{1} \and
Nils Kemmerzell\inst{1}\and
Mohamed Abdelaal\inst{2}\and 
Davor Stjelja \inst{3} }
\authorrunning{T. Clement et al.}
\institute{Friedrich-Alexander-University Erlangen-Nürnberg, Lange Gasse 20, 90403 Nürnberg, Germany \and
Software AG, Uhlandstraße 12, 64297 Darmstadt, Germany \and Granlund, Malminkaari 21, 00700 Helsinki, Finland}
\maketitle              % typeset the header of the contribution
\vspace{-2pt}
\begin{abstract}
This paper presents an approach integrating explainable artificial intelligence (XAI) techniques with adaptive learning to enhance energy consumption prediction models, with a focus on handling data distribution shifts. Leveraging SHAP clustering, our method provides interpretable explanations for model predictions and uses these insights to adaptively refine the model, balancing model complexity with predictive performance.
We introduce a three-stage process: (1) obtaining SHAP values to explain model predictions, (2) clustering SHAP values to identify distinct patterns and outliers, and (3) refining the model based on the derived SHAP clustering characteristics. Our approach mitigates overfitting and ensures robustness in handling data distribution shifts.
We evaluate our method on a comprehensive dataset comprising energy consumption records of buildings, as well as two additional datasets to assess the transferability of our approach to other domains, regression, and classification problems. Our experiments demonstrate the effectiveness of our approach in both task types, resulting in improved predictive performance and interpretable model explanations.
\end{abstract}

\section{Introduction}

Buildings contribute significantly to global energy consumption and $\text{CO}^2$ emissions, offering substantial leverage to address the climate crisis. The building sector accounts for 30-40\% of global energy consumption and 30-50\% of human greenhouse gas emissions~\cite{skillington2022review,azar2011decision}. Accurate building energy consumption prediction is vital for reducing energy use and emissions by informing decision-making, evaluating design options, and optimizing demand-supply management. However, prediction is complex due to factors like building characteristics, equipment, weather conditions, and occupants' habits~\cite{kwok2011study,zambrano2021towards,carlucci2021impact,azar2020simulation}.
Data-driven approaches are essential for predicting building energy consumption~\cite{amasyali2018review} but face challenges: (1) models may underperform outside their training range, requiring representative training data, and (2) the inherent opacity of many models restricts a clear understanding of their internal workings.

In light of these challenges, Explainable AI (XAI) emerges as a beacon, illuminating the intricate workings of AI models and offering insights understandable to humans~\cite{WEBER2023154}. Techniques like SHapley Additive exPlanations (SHAP)~\cite{NIPS2017_7062} delve into the contributions of individual features to a model's predictions, serving as valuable tools for model refinement. Merging these insights with adaptive learning strategies holds promise in forging resilient models that can adeptly navigate the treacherous waters of data distribution shifts, ensuring consistent prediction accuracy even amidst ever-evolving conditions.

Explainable AI (XAI) addresses these limitations by offering transparent and interpretable AI systems, enabling model improvement through human-friendly explanations~\cite{WEBER2023154}.

In this paper, we tackle the challenges of predicting building energy consumption, especially when faced with unforeseen future data or abrupt changes in consumption patterns, such as those observed during and after the COVID-19 pandemic. To address these challenges, we develop an energy consumption predictor utilizing Explainable AI (XAI) techniques, specifically SHapley Additive exPlanations (SHAP)~\cite{NIPS2017_7062}, in conjunction with automated hyperparameter tuning (AHT) algorithms. By analyzing the impact of various features on the model's predictions, we introduce a novel SHAP Clustering-based Adaptive Learning (SCAL) approach that integrates XAI, dimensional reduction, and clustering methods to enhance model performance on test sets experiencing data shifts from the training set. In summary, this paper offers the following contributions:

\begin{itemize}
\item Address challenges in data-driven energy consumption prediction, specifically for unforeseen data and sudden consumption pattern changes.
\item Introduce the novel SHAP Clustering-based Adaptive Learning (SCAL) approach, leveraging XAI techniques, dimensional reduction, and clustering to improve model performance on test sets experiencing data shifts.
\item Demonstrate SCAL's effectiveness and versatility across regression and classification problems and diverse datasets, highlighting its potential for enhancing model robustness and accuracy in various domains.
\end{itemize}

\section{Related Work}
In this section, we review key areas related to our study: predicting energy use, improving models with XAI, and the role of SHAP. 
This encompasses an exploration of machine learning's role in energy consumption prediction, the potential of XAI in model improvement, and the rationale for selecting SHAP within our methodology.
\subsection{Energy consumption prediction}
Energy consumption prediction is essential for energy efficiency and sustainability. Numerous studies have explored various ML techniques based on historical data to predict future energy consumption. 
These techniques optimize energy usage, reduce carbon emissions, and lead to cost savings and environmental benefits. For instance, XGBoost, deep learning, and Deep Reinforcement Learning have been employed in different settings~\cite{shen2023machine,amiri2021peeking,ZHANG2023125468,olu2022building,KHAN2023112705,JIN2023105458}. These studies highlight the importance of energy consumption prediction and illustrate the diverse ML techniques proposed to tackle the problem. Distinctively, our method harnesses XAI techniques to predict energy consumption patterns with greater accuracy and interpretability, even in scenarios of sudden data shifts like those prompted by the COVID-19 pandemic.

\subsection{XAI-based model improvement}
XAI-based model improvements aim to enhance various properties of a model, including performance, convergence, robustness, efficiency, reasoning, and equality~\cite{WEBER2023154}. Several approaches have been proposed to improve models using XAI. For instance, explanation-guided dropout~\cite{zunino2021excitation} reduces overfitting and encourages alternative solutions. 
At the same time, eXplanatory Interactive Learning (XIL)~\cite{teso2019explanatory}, Prototypical Relevance Propagation~\cite{gautam2023looks}, and Guided Zoom~\cite{bargal2018guided,bargal2021guided} augment various aspects of the model, such as data, intermediate features, loss, gradient, and the model itself. Moreover, human-in-the-loop knowledge and interaction have been employed to improve model performance and reasoning~\cite{zaidan2007using,zaidan2008modeling,hendricks2018women,zhang2016rationale,mcdonnell2016relevant,tong2001support,judah2012active,shivaswamy2015coactive,gal2017deep}.
Our approach innovatively combines XAI techniques with clustering and adaptive learning, introducing a systematic method for improving model performance and interpretability, especially in the presence of data shifts.

\subsection{SHapley Additive exPlanations (SHAP)}
SHAP is a prominent XAI approach that provides insights into the contribution of individual features to a model's predictions~\cite{NIPS2017_7062}. 
SHAP values, derived from Shapley values in cooperative game theory, have several desirable properties, including consistency, local accuracy, and missingness consistency~\cite{hart1989shapley,nguyen2021evaluation}. 
While extensions like SHAP clustering have been applied to classification problems to identify and characterize similar instance clusters~\cite{durvasula2022characterizing}, their use in regression problems has been limited. Our study pioneers the application of SHAP clustering to regression through the SCAL framework, marking a significant contribution to the field.

The choice to incorporate SHAP as the backbone of our SCAL framework stems from its distinctive features that set it apart in the landscape of XAI methods. With SHAP, we can obtain an accurate and balanced depiction of how each feature influences every single prediction, and importantly, these influences maintain stability when aggregated across all predictions.
Another aspect of SHAP is its adept handling of missing values, a prevalent issue in real-world datasets. This ability to accurately account for the effect of absent data is crucial, ensuring the robustness of our analysis.
Moreover, SHAP's competence in offering explanations at the level of individual data instances aligns perfectly with the goal of our SCAL framework - to learn and adapt in response to data changes. This characteristic enables our framework to dynamically adjust model parameters in light of shifts in individual instance explanations.

\section{Methodology}

\begin{figure*}[h!]
    \centering
    \includegraphics[width=0.99\linewidth]{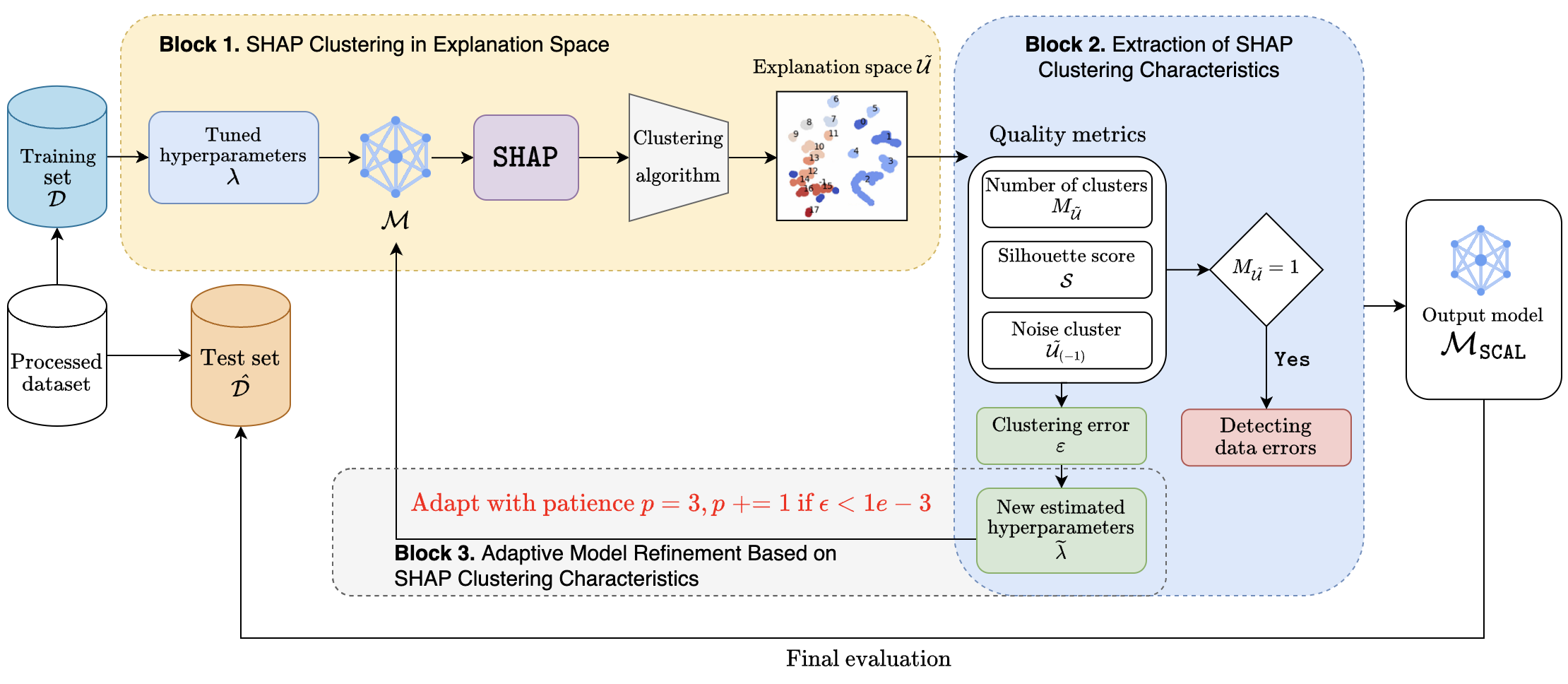}
    \caption{SCAL Pipeline: Adaptive Learning via SHAP Clustering in Three Building Blocks.}
    \label{fig:process}
    \vspace{-20pt}
\end{figure*}

The SHAP Clustering-based Adaptive Learning (SCAL) framework is designed to facilitate model adaptability in response to data shifts, enhancing both model interpretability and performance. This is achieved through the integration of three interconnected building blocks. The full pipeline, visually represented in Fig.~\ref{fig:process}, comprises the following key stages:
\textit{Building Block 1: SHAP Clustering} initiates the framework by clustering instances based on their explanation similarity, using SHAP values within an embedding space (explanation space). This step identifies distinct explanation characteristics that can inform model adaptation.
\textit{Building Block 2: Extraction of SHAP Clustering Characteristics} builds upon the clusters formed in Block 1, this block characterizes the clustering using three quality metrics. The insights derived are instrumental in understanding the relationships between instances and the behavior of the model, providing critical information for model adaptation.
\textit{Building Block 3: Model Adaptation} utilizes the SHAP clustering characteristics derived in Block 2 to adapt and fine-tune the model, enhancing its ability to adapt to data shifts. This process not only improves the model's performance but also enhances its interpretability, completing the SCAL framework.
Each building block within the SCAL framework has a distinct and crucial role, collectively contributing to the framework's adaptability and effectiveness. The dynamic interplay of these building blocks allows the model to learn and adapt as it encounters shifts in data, making it robust and reliable. The following subsections will delve into each building block in more depth, elucidating the specific roles they play and the methodologies they employ within the framework. For a visual overview of how these building blocks interact to form the complete SCAL pipeline, refer to Fig.~\ref{fig:process}.

\subsection{Building Block 1: SHAP Clustering in Explanation Space}
The first building block in the SCAL framework is SHAP Clustering in Explanation Space. This building block forms the foundation for the entire framework, identifying and characterizing groups of instances based on explanation similarity and enabling subsequent building blocks to extract valuable insights and adapt the model.
We use SHAP values, or SHapley Additive exPlanations, to understand each feature's contribution to an instance's prediction. By representing each instance $x$ with its corresponding SHAP values, we transform the original input space into an explanation space (embedding space), allowing us to focus on the model's reasoning instead of its raw predictions.
In the explanation space, we use DBSCAN \cite{DBSCAN} to cluster instances based on SHAP value similarities. Clustering instances with similar explanation patterns identifies subgroups sharing common model behaviors and feature interactions, simplifying the analysis and interpretation of the model's decisions. Moreover, this clustering process uncovers the underlying structure or trends in the explanation space, providing valuable insights into the model's behavior.
To implement this building block, we follow these steps:
\begin{enumerate}
\item Train a supervised model using AHT algorithms for each building.
\item Compute SHAP values for instances with the AHT-tuned trained model.
\item Apply UMAP to extract a 2D space of SHAP values.
\item Utilize DBSCAN for creating SHAP clusters embedding.
\end{enumerate}

By completing this building block, we have established a solid groundwork in the explanation space, with well-defined clusters that pave the way for the following building blocks to unlock further insights and adapt the model, resulting in superior performance and better understanding.

\subsection{Building Block 2: Extraction of SHAP Clustering Characteristics}
The second building block in the SCAL framework extracts valuable characteristics from the SHAP-based clusters created in Building Block 1. These insights into the model's behavior within clusters pave the way for targeted adaptation in Building Block 3.
After clustering instances in the explanation space using SHAP values, we aim to understand the model's behavior within these subgroups. To this end, we use three quality metrics to describe the explanation space (as shown in Fig.~\ref{fig:quality_metrics.png}):

\begin{SCfigure}
    \centering
    \includegraphics[width=0.4\linewidth]{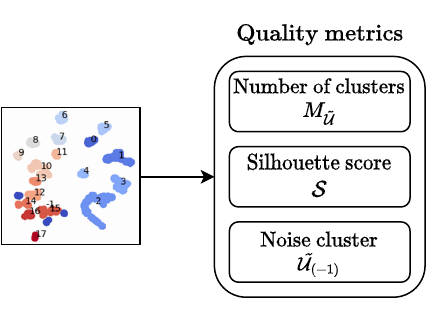}
    \vspace{-10pt}
    \caption{Overview of the three quality metrics: number of clusters ($M$), intra-inter distances between clusters, and the presence of noise cluster.}
    \label{fig:quality_metrics.png}
    \vspace{-10pt}
\end{SCfigure}

\begin{itemize}
\item \textbf{Number of clusters} ($M$): Each cluster reflects a set of logical rules learned by the model. More clusters suggest greater information extraction from the data. The optimal number is estimated using silhouette analysis~\cite{shutaywi2021silhouette}.
\item \textbf{Silhouette score}: Reflects cluster separation and cohesion. This score uses intra-inter distances between clusters to estimate optimal cluster count~\cite{shutaywi2021silhouette}. Distinct clusters can lead to improved anomaly detection as these data points lie more likely between classified clusters.
\item \textbf{Presence of noise cluster}:  Aids in identifying well-defined clusters in noisy data and enhances anomaly detection~\cite{dave1991characterization,bigdeli2017fast}.
Noise clusters can be helpful to reduce overfitting~\cite{goodfellow2016deep}. DBSCAN labels noise cluster as $\mathcal{~{U}}_{(-1)}$~\cite{DBSCAN}.
\end{itemize}

We apply silhouette analysis~\cite{shutaywi2021silhouette} to estimate the optimal number of clusters, accounting for the trade-off between bias and variance, sample size, and computational cost by taking the intra-inter distances between clusters into consideration~\cite{still2004many}.
Then, the silhouette score for a cluster space is computed as the mean of all silhouette scores for each sample.
In the SCAL framework, we also use SkopeRules~\cite{scikitle83:online} to generate interpretable decision rules for each SHAP-based cluster. This approach offers both interpretability and flexibility in rule generation, making it suitable for feature selection and model explanation. By applying SkopeRules with a one-vs-all approach, we can learn decision rules for each cluster, gaining insights into the relationships between instances and the underlying model behavior.

\subsection{Building Block 3: Adaptive Model Refinement Based on SHAP Clustering Characteristics}
In this stage of the SCAL method, we leverage the information obtained from SHAP clustering to iteratively refine the model by complying with defined conditions extracted from SHAP clustering characteristics. The objective is to adapt the model's hyperparameters based on the quality metrics derived from the SHAP clusters embedding in an unsupervised way, so as to avoid overfitting to the training dataset. The pseudo-code for the SCAL algorithm is shown in Algorithm~\ref{algorithm:process}.
The adaptive refinement process starts by initializing the hyperparameters, $\bar{\lambda}$, to the current best-fit values, $\lambda$, found using Automated Hyperparameter Tuning (AHT). The maximum depth of the model, $\mathtt{max\_depth}$, is reduced to create a shallower tree, which helps mitigate overfitting. The refined model is then evaluated by comparing its performance against the previous model using silhouette score improvements, calculated using the intra-inter distances between clusters and noise cluster reduction.
If the refined model shows a significant improvement over the previous model, the model is updated, and the regularization hyperparameter, $\mathtt{gamma}$, is increased. This further enhances the model's ability to generalize, making learning complex relationships between features and the target variable more difficult. If the refined model does not significantly improve, the maximum depth is further reduced, and the process is repeated.
This adaptive model refinement process is carried out iteratively, with a predefined number of patience $\rho$ to ensure convergence. We choose $\rho$ as 3 since the model can be adapted via one step of depth reduction and one step of gamma reduction. The $\Gamma$ is set as 0.01 as we expect the silhouette score improvement should be higher than the greedy noise exploration. Once the patience value is reached, the best model obtained throughout the iterations is set as the final SCAL. This approach allows the model to adapt dynamically to the data, incorporating the insights derived from SHAP clustering to balance bias and variance better.
Building Block 3 represents an innovative approach to model refinement, where the adaptive learning process is guided by the insights gained from SHAP clustering. This not only helps reduce the model's bias and prevent overfitting but also ensures that the final model is robust and capable of effectively handling anomalous data patterns.

\vspace{-10pt}
\providecommand{\SetAlgoLined}{\SetLine}
\providecommand{\DontPrintSemicolon}{\dontprintsemicolon}

\begin{algorithm}[h!]

\scriptsize
\SetAlgoLined
\SetArgSty{textup}
\DontPrintSemicolon
\KwData{Training set of a building $\mathcal{D}=(x,y)$}
\KwOut{SCAL model $\mathcal{M}_{\texttt{SCAL}}$}
\Begin{
$p \leftarrow 0$ \tcp*[r]{Patience}
% $\mathcal{U} \leftarrow \texttt{DBSCAN}(\texttt{UMAP}(x))$ \tcp*[r]{Data space}
$\lambda \leftarrow \texttt{XGBTune}(\mathcal{D})$ \tcp*[r]{AHT on the training set}
% $\mathcal{\tilde{N}} \leftarrow max(s(y_i))$, $s(y_i) \leftarrow \frac{b(y_i)-a(y_i)}{max\{a(y_i),b(y_i)\}}$, $\mathcal{C}\leftarrow(\tilde{N}, s_{(-1)})$\;

$\mathcal{M} \leftarrow \texttt{XGBRegressor}(\lambda)$\ \tcp*[r]{Model with tuned hyperparameters}
$\tilde{y} \leftarrow \mathcal{M}(\mathcal{D})$ \tcp*[r]{Model predictions on training set}
$\mathcal{\tilde{U}} \leftarrow \texttt{DBSCAN}(\texttt{UMAP}(\texttt{SHAP}(\tilde{y})))$ \tcp*[r]{Explanation space}
$\mathcal{S} \leftarrow \texttt{silhouette\_score}(\mathcal{\tilde{U}})$ \tcp*[r]{Silhouette score computation}
$\bar{\lambda} \leftarrow \lambda$ \tcp*[r]{Initialize adaptive hyperparameters}
$\bar{\lambda}(\texttt{max\_depth}) \leftarrow \bar{\lambda}(\texttt{max\_depth})-1 $ \tcp*[r]{Reduce the depth}
% Model improvement part
\While{$p < \rho$}{
\lIf(\tcp*[f]{Check value}){\textsf{$\bar{\lambda}(\texttt{max\_depth}$}) $<$ 1}{break}
$\mathcal{M}_{p} \leftarrow \texttt{XGBRegressor}(\bar{\lambda})$\ \tcp*[r]{Model with adaptive hyperparameters}
$\tilde{y}' \leftarrow \mathcal{M}_{p}(\mathcal{D})$ \;
$\mathcal{\tilde{U}'} \leftarrow \texttt{DBSCAN}(\texttt{UMAP}(\texttt{SHAP}(\tilde{y}')))$ \tcp*[r]{Explanation space}
$\mathcal{S}' \leftarrow \texttt{silhouette\_score}(\mathcal{\tilde{U}'})$\;
$\nabla \mathcal{S} = \mathcal{S}' - \mathcal{S} $ \tcp*[r]{Silhouette score improvement}

$\nabla \tilde{\mathcal{U}}_{(-1)} = \neg\texttt{bool}(\tilde{\mathcal{U}}_{(-1)}) * \Gamma$ \tcp*[r]{Noise cluster}
$\epsilon \leftarrow \nabla \mathcal{S} + \nabla \tilde{\mathcal{U}}_{(-1)}$ \tcp*[r]{Loss}

\eIf(\tcp*[f]{Check the loss with the threshold}){$\epsilon >= 1e-3$}{ 
    $\mathcal{M} \leftarrow \mathcal{M}_{p}$ \tcp*[r]{Update the best model}
    $\mathcal{S} \leftarrow \mathcal{S}'$ \tcp*[r]{Update the best silhouette score}
    \eIf(\tcp*[f]{Update gamma}){\textsf{$\bar{\lambda}(\texttt{gamma}$}) == 0}{
        $\bar{\lambda}(\texttt{gamma}) \leftarrow 0.001$
    }{
        $\bar{\lambda}(\texttt{gamma}) \leftarrow \bar{\lambda}(\texttt{gamma})*10 $ 
    } 
 
    $p \leftarrow 0$ \tcp*[r]{Reset the patience}
}{
    $\bar{\lambda}(\texttt{max\_depth}) \leftarrow \bar{\lambda}(\texttt{max\_depth})-1 $ \tcp*[r]{Reduce the depth}
     $\bar{\lambda}(\texttt{gamma}) \leftarrow \lambda(\texttt{gamma}) $ \tcp*[r]{Reset to the tuned gamma}
    $p \leftarrow p+1$ \tcp*[r]{Increase the patience}

}
}
$\mathcal{M}_{\texttt{SCAL}} \leftarrow \mathcal{M}$ \tcp*[r]{SCAL model as the best model}
\Return $\mathcal{M}_{\texttt{SCAL}}$\;
}
\caption{SCAL procedure}\label{algorithm:process}

\end{algorithm}
%\vspace{-30pt}

\section{Experimental Setup and Data Set}
\begin{SCfigure}
\centering
\scalebox{0.45}{\input{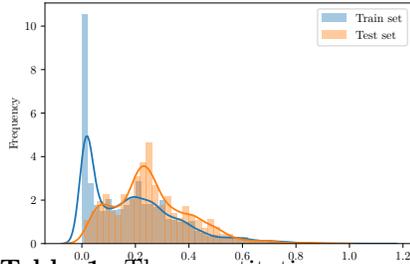}}
\vspace{-20pt}
\caption{The distribution of energy consumption of a general educational institution's training and test set (building 1) reflects the significant growth in energy consumption after the COVID-19 pandemic.}
\label{fig:dist_1}
\vspace{-15pt}
\end{SCfigure}

In our experimental setup, we apply the SHAP Clustering-based Adaptive Learning (SCAL) framework to the specific use case of predicting energy consumption in buildings. This use case bears significant relevance in the context of optimizing energy usage, reducing operating costs, and minimizing environmental impacts. Accurate predictions can aid in the formulation and implementation of efficient energy-saving strategies and policies.
Our experiments employ energy consumption data from 36 buildings collected between January 1st, 2019, and June 1st, 2022. The data, recorded in MWh, includes features such as the purpose of use, outside temperature, area, and volume of the buildings. The buildings serve various purposes, including General education, University, Research institutes, and Offices.
To ensure data integrity, we preprocess the data by replacing negative consumption values and applying linear interpolation for missing values. We divide the data, setting all data before 2022 as the training set and data from 2022 onwards as the test set. As energy consumption is serialized hourly, this task constitutes a time series prediction.
Illustratively, consider a general educational institution (building 1). As shown in Fig.~\ref{fig:dist_1}, the training and test set distributions exhibit notable differences owing to the increased energy consumption following the COVID-19 pandemic.

\section{Results}
\subsection{SCAL Performance}

\newcommand{\false}{\textcolor{red}{\ding{55}}}
\newcommand{\true}{\textcolor{teal}{\ding{51}}}
\begin{table}[t!]
\setlength{\tabcolsep}{5pt}
\centering\small
\caption{The quantitative comparison between the AHT and SCAL on the Energy consumption data set. The best results are in bold. We only consider the silhouette score (SS) in the training set. The arrow indicates that the lower/higher, the better. Noise cluster presents ($\true$) and does not present ($\false$).}
\resizebox{\linewidth}{!}{%
\begin{tabular}{cccccccccc} 
\toprule
\multirow{2}{*}[-2pt]{Building ID} & \multirow{2}{*}[-2pt]{Data set} & \multicolumn{4}{c}{AHT} & \multicolumn{4}{c}{SCAL}  \\
\cmidrule(lr){3-6} \cmidrule(lr){7-10}
 &  & $\tilde{\mathcal{U}}_{(-1)}$ & $SS \uparrow$ & $RMSE \downarrow$ & $r^2 \uparrow$ & $\tilde{\mathcal{U}}_{(-1)}$ &$SS \uparrow$ & $RMSE \downarrow$ & $r^2 \uparrow$ \\
\midrule
\multirow{2}{*}[-2pt]{1} & Training  & $\false$ & 0.68514 & 0.05622 & 0.99684  & $\true$ & \textbf{0.68721} & 0.06783 & 0.99540  \\
\cmidrule{2-10}
& Test & - & - & 0.20527 & 0.94403  & - & -  & \textbf{0.20144} & \textbf{0.94610} \\

\midrule
\multirow{2}{*}[-2pt]{2} & Training & $\false$ & 0.71530 & 0.23016 & 0.94689 & $\false$ & \textbf{0.72695} & 0.49800 & 0.75136 \\
\cmidrule{2-10}
& Test & - & - &  0.51697 & 0.48786  & - & - & \textbf{0.50698} & \textbf{0.50743} \\

\midrule
\multirow{2}{*}[-2pt]{3} & Training & $\true$ & 0.77496 & 0.030791 & 0.90516 & $\true$ & \textbf{0.86414} & 0.46356 & 0.78504 \\
\cmidrule{2-10}
& Test & - & - & 0.58662 & 0.62682 & - & - & \textbf{0.56799} & \textbf{0.65014} \\

\midrule
\multirow{2}{*}[-2pt]{4} & Training & $\true$ & 0.21900 & 0.05509 & 0.99697 & $\true$ & \textbf{0.25013} & 0.07266 & 0.99452 \\
\cmidrule{2-10}
& Test & - & - & 0.15288 & 0.97033 & -& - & \textbf{0.14883} & \textbf{0.97188} \\
\bottomrule
\end{tabular}
}
\label{tab:quantitative_results}
\end{table}

We compare SCAL's performance with the traditional AHT (Adaptive Hyperparameter Tuning) to highlight SCAL's superior model adaptation and accuracy. AHT optimizes the model using hyperparameter tuning without considering the explanation space. In contrast, SCAL leverages SHAP clustering characteristics for model adaptation. As demonstrated in Table~\ref{tab:quantitative_results}, SCAL outperforms AHT on the test sets, yielding lower RMSE and higher $r^2$ values, indicating a stronger generalization to unseen data. Even with a slightly worse training set fit, SCAL provides a better silhouette score in the explanation space, illustrating the effective use of SHAP clustering characteristics. Importantly, SCAL offers valuable insights into the explanation space that enhance understanding of model performance and inform further model refinement. Unlike AHT, this additional information increases SCAL's potential for improvement and adaptability. Thus, SCAL's performance underscores the advantage of a model that adapts based on SHAP clustering characteristics, delivering improved generalization and valuable insights for model adaptation.

\subsection{Cluster Analysis in Explanation Space}
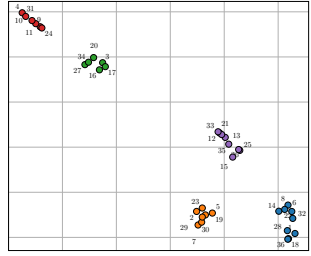
\begin{SCfigure}
    \centering
    \resizebox{.35\linewidth}{!}{%
        % This file was created with tikzplotlib v0.10.1.
\begin{tikzpicture}

\definecolor{crimson2143940}{RGB}{214,39,40}
\definecolor{darkgray176}{RGB}{176,176,176}
\definecolor{darkorange25512714}{RGB}{255,127,14}
\definecolor{forestgreen4416044}{RGB}{44,160,44}
\definecolor{mediumpurple148103189}{RGB}{148,103,189}
\definecolor{steelblue31119180}{RGB}{31,119,180}

\begin{axis}[
tick align=inside,
tick pos=left,
x grid style={darkgray176},
xmajorgrids,
xmin=-3.99855047464371, xmax=7.13039380311966,
xtick style={draw=none},
y grid style={darkgray176},
ymajorgrids,
ymin=-0.582565124705434, ymax=10.4601057056338,
ytick style={draw=none},
yticklabels=\empty,
xticklabels=\empty,
]
\addplot [draw=black, fill=steelblue31119180, mark=*, only marks]
table{%
x  y
6.62453269958496 0.173173904418945
6.50576591491699 1.1492213010788
6.36507558822632 1.43245220184326
6.02370309829712 1.16019022464752
6.38342142105103 -0.0569759197533131
6.24384832382202 1.24128317832947
6.34152269363403 0.300188332796097
6.54127836227417 0.84234744310379
6.35849142074585 -0.0806255415081978
};
\addplot [draw=black, fill=darkorange25512714, mark=*, only marks]
table{%
x  y
2.97494673728943 1.15051651000977
3.55966758728027 1.07868695259094
3.03077530860901 0.549641311168671
3.30142021179199 0.996193051338196
3.18672776222229 1.30289649963379
3.16307950019836 0.675048172473907
3.18650913238525 0.901816189289093
};
\addplot [draw=black, fill=forestgreen4416044, mark=*, only marks]
table{%
x  y
-0.509819984436035 7.74483823776245
-0.628013968467712 7.42748022079468
-0.40979927778244 7.56949758529663
-0.84404981136322 7.9712233543396
-1.16587376594543 7.65859317779541
-1.03352785110474 7.76590824127197
};
\addplot [draw=black, fill=crimson2143940, mark=*, only marks]
table{%
x  y
-3.49268937110901 9.95816612243652
-2.82496166229248 9.33763790130615
-3.36314415931702 9.79606819152832
-2.98969388008118 9.47526836395264
-2.7630615234375 9.28439521789551
-3.12834095954895 9.61230373382568
};
\addplot [draw=black, fill=mediumpurple148103189, mark=*, only marks]
table{%
x  y
3.82911014556885 4.62955760955811
4.03719329833984 4.43312501907349
4.31379413604736 3.56029629707336
3.90682911872864 4.56000328063965
4.57638502120972 3.85738945007324
4.53534650802612 3.8950297832489
3.77231740951538 4.68192625045776
4.16733884811401 4.13279104232788
};
\draw (axis cs:6.61955522264056,0.424143696017797) node[
  scale=0.5,
  anchor=east,
  text=black,
  rotate=0.0
]{1};
\draw (axis cs:2.96996844190309,1.10868821141) node[
  scale=0.5,
  anchor=north east,
  text=black,
  rotate=0.0
]{2};
\draw (axis cs:-0.504841960093871,7.78666653636222) node[
  scale=0.5,
  anchor=south west,
  text=black,
  rotate=0.0
]{3};
\draw (axis cs:-3.49766739545117,9.99999442103629) node[
  scale=0.5,
  anchor=south east,
  text=black,
  rotate=0.0
]{4};
\draw (axis cs:3.60335449126755,1.13038514486719) node[
  scale=0.5,
  anchor=south west,
  text=black,
  rotate=0.0
]{5};
\draw (axis cs:6.43217913376827,1.33129712793173) node[
  scale=0.5,
  anchor=south west,
  text=black,
  rotate=0.0
]{6};
\draw (axis cs:3.05077512407337,0.0618311350969911) node[
  scale=0.5,
  anchor=north east,
  text=black,
  rotate=0.0
]{7};
\draw (axis cs:6.36009750967531,1.47428050044332) node[
  scale=0.5,
  anchor=south east,
  text=black,
  rotate=0.0
]{8};
\draw (axis cs:-2.8661507680987,9.41719433856695) node[
  scale=0.5,
  anchor=south,
  text=black,
  rotate=0.0
]{9};
\draw (axis cs:-3.3798748452481,9.82389770043737) node[
  scale=0.5,
  anchor=north east,
  text=black,
  rotate=0.0
]{10};
\draw (axis cs:-2.98509173791618,9.29366813892111) node[
  scale=0.5,
  anchor=north east,
  text=black,
  rotate=0.0
]{11};
\draw (axis cs:3.7839588283666,4.62568225362931) node[
  scale=0.5,
  anchor=north east,
  text=black,
  rotate=0.0
]{12};
\draw (axis cs:4.21320760865638,4.51096693481787) node[
  scale=0.5,
  anchor=west,
  text=black,
  rotate=0.0
]{13};
\draw (axis cs:6.01374694119511,1.20201852324757) node[
  scale=0.5,
  anchor=south east,
  text=black,
  rotate=0.0
]{14};
\draw (axis cs:4.23701961541737,3.349610786218) node[
  scale=0.5,
  anchor=north east,
  text=black,
  rotate=0.0
]{15};
\draw (axis cs:-0.637970559240408,7.38565192219491) node[
  scale=0.5,
  anchor=north east,
  text=black,
  rotate=0.0
]{16};
\draw (axis cs:-0.399842687009747,7.52766928669686) node[
  scale=0.5,
  anchor=north west,
  text=black,
  rotate=0.0
]{17};
\draw (axis cs:6.3933780118237,-0.0988042183530822) node[
  scale=0.5,
  anchor=north west,
  text=black,
  rotate=0.0
]{18};
\draw (axis cs:3.56962417805296,1.03685865399117) node[
  scale=0.5,
  anchor=north west,
  text=black,
  rotate=0.0
]{19};
\draw (axis cs:-0.824894428691564,8.28424963053273) node[
  scale=0.5,
  anchor=south,
  text=black,
  rotate=0.0
]{20};
\draw (axis cs:3.79503085242629,4.80376502501626) node[
  scale=0.5,
  anchor=south west,
  text=black,
  rotate=0.0
]{21};
\draw (axis cs:6.36060139826218,1.20310220330333) node[
  scale=0.5,
  anchor=north,
  text=black,
  rotate=0.0
]{22};
\draw (axis cs:3.17677160512028,1.34472479823384) node[
  scale=0.5,
  anchor=south east,
  text=black,
  rotate=0.0
]{23};
\draw (axis cs:-2.75115764464898,9.24167298073104) node[
  scale=0.5,
  anchor=north west,
  text=black,
  rotate=0.0
]{24};
\draw (axis cs:4.63211727493148,3.89820336226507) node[
  scale=0.5,
  anchor=south west,
  text=black,
  rotate=0.0
]{25};
\draw (axis cs:4.40547908635219,3.89639007259781) node[
  scale=0.5,
  anchor=north,
  text=black,
  rotate=0.0
]{26};
\draw (axis cs:-1.19833689795744,7.62505461568168) node[
  scale=0.5,
  anchor=north east,
  text=black,
  rotate=0.0
]{27};
\draw (axis cs:6.22858163448968,0.266057470753733) node[
  scale=0.5,
  anchor=south east,
  text=black,
  rotate=0.0
]{28};
\draw (axis cs:2.75590174029012,0.188465642215068) node[
  scale=0.5,
  anchor=south east,
  text=black,
  rotate=0.0
]{29};
\draw (axis cs:3.06570969310727,0.563770718294248) node[
  scale=0.5,
  anchor=north west,
  text=black,
  rotate=0.0
]{30};
\draw (axis cs:-3.17978337743864,9.91913392176419) node[
  scale=0.5,
  anchor=south,
  text=black,
  rotate=0.0
]{31};
\draw (axis cs:6.6462444952534,1.03849903773333) node[
  scale=0.5,
  anchor=west,
  text=black,
  rotate=0.0
]{32};
\draw (axis cs:3.73418768372086,4.71066238883263) node[
  scale=0.5,
  anchor=south east,
  text=black,
  rotate=0.0
]{33};
\draw (axis cs:-1.04393942569089,7.7823100473355) node[
  scale=0.5,
  anchor=south east,
  text=black,
  rotate=0.0
]{34};
\draw (axis cs:4.15738225734132,4.09096274372811) node[
  scale=0.5,
  anchor=north east,
  text=black,
  rotate=0.0
]{35};
\draw (axis cs:6.34690680189885,-0.101539690807838) node[
  scale=0.5,
  anchor=north east,
  text=black,
  rotate=0.0
]{36};
\end{axis}

\end{tikzpicture}
    }
    \vspace{-5pt}
    \caption{The categories of buildings based on their similarity in the explanation space. Five buildings categories are clustered.}
    \label{fig:cluster_building}
    \vspace{-10pt}
\end{SCfigure}

In this section, we perform the cluster analysis in the explanation space by comparing the performance of the SCAL on the energy consumption test set with anomalous data points to that of the initial AHT model to observe the model improvement. 
At first, to identify similarities between buildings, we cluster all buildings based on their mean SHAP values, which provides insight into their explanation space. Fig.~\ref{fig:cluster_building} illustrates five distinct groups of buildings that exhibit high resemblance in their explanation spaces.
To better understand the clustering behavior in the explanation space, we choose representative buildings from each group for further analysis. We select buildings with ID 1, 2, 3, 4 (model improvement cases), and 12 (data error detection case) to represent the five building groups.

\subsubsection{Model improvement cases}
Fig.~\ref{fig:building_cluster_2} presents the SHAP clusters embedding on the training set of buildings 1, 2, 3, and 4 for both AHT and SCALs. The SCAL is expected to show a noise cluster, indicated by index -1, while the AHT model does not (Building 1, Fig.~\ref{fig:b1}). Additionally, the number of clusters in the SCAL is higher than in the AHT model, implying that the SCAL can learn more rules to explain each data point. This results in improved model performance on the test set, as demonstrated in Table~\ref{tab:quantitative_results}.
However, the number of clusters alone is not a consistent metric for assessing cluster quality. Therefore, we employ the silhouette score as the primary metric for evaluating the quality of clusters in the explanation space.

\newcommand{\rulesep}{\unskip\ \vrule\ }
\begin{figure}[h]
    
    \centering
    \captionsetup[subfigure]{justification=centering}
    \begin{subfigure}[b]{0.24\linewidth}
        \centering
        \includegraphics[width=\textwidth]{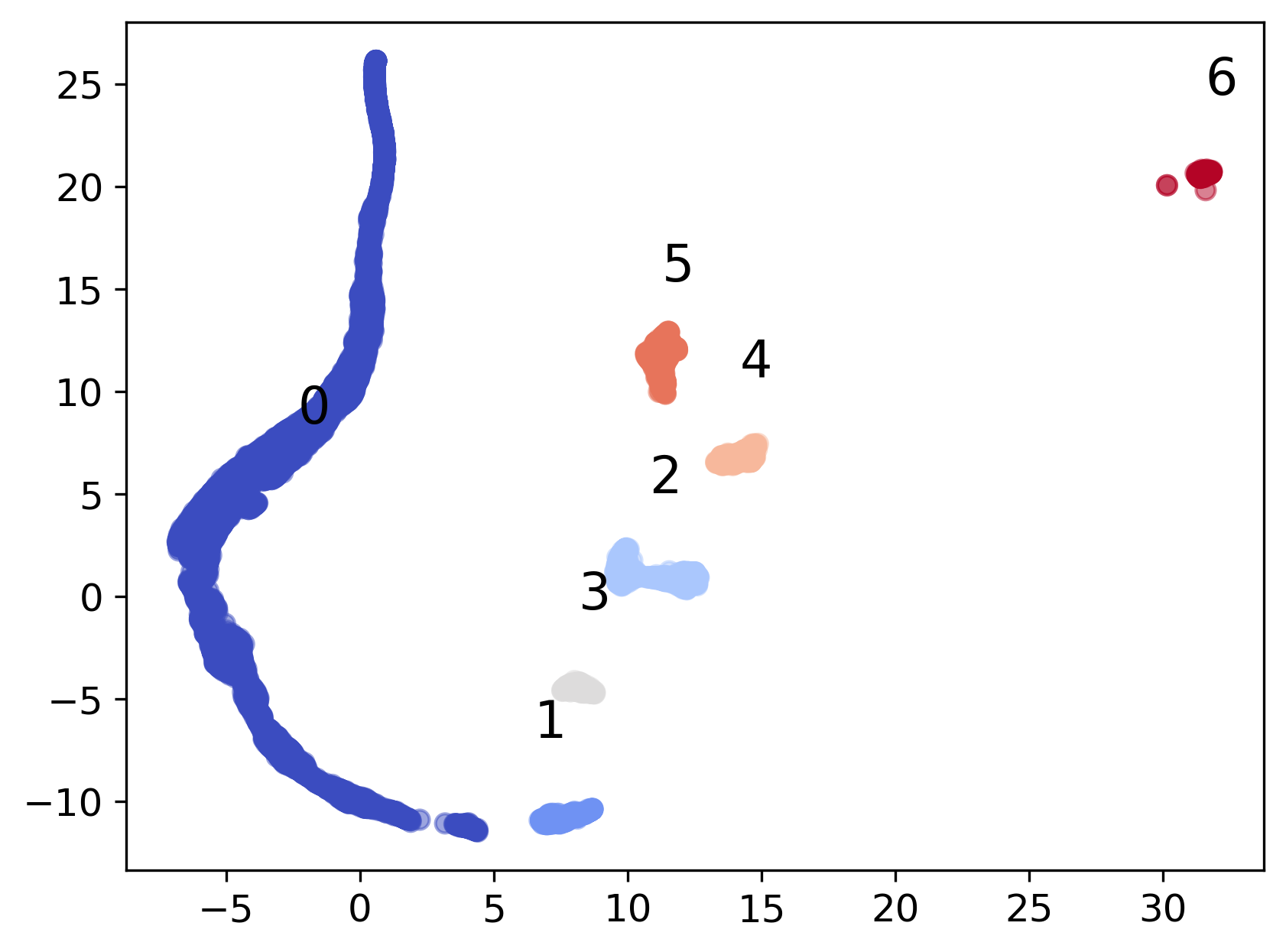}
        \caption{AHT \\(Building 1) \\ $\tilde{\mathcal{U}}_{(-1)}$=\false\\SS=0.68514}
        \label{fig:b1}
    \end{subfigure}
    \begin{subfigure}[b]{0.24\linewidth}
        \centering
        \includegraphics[width=\textwidth]{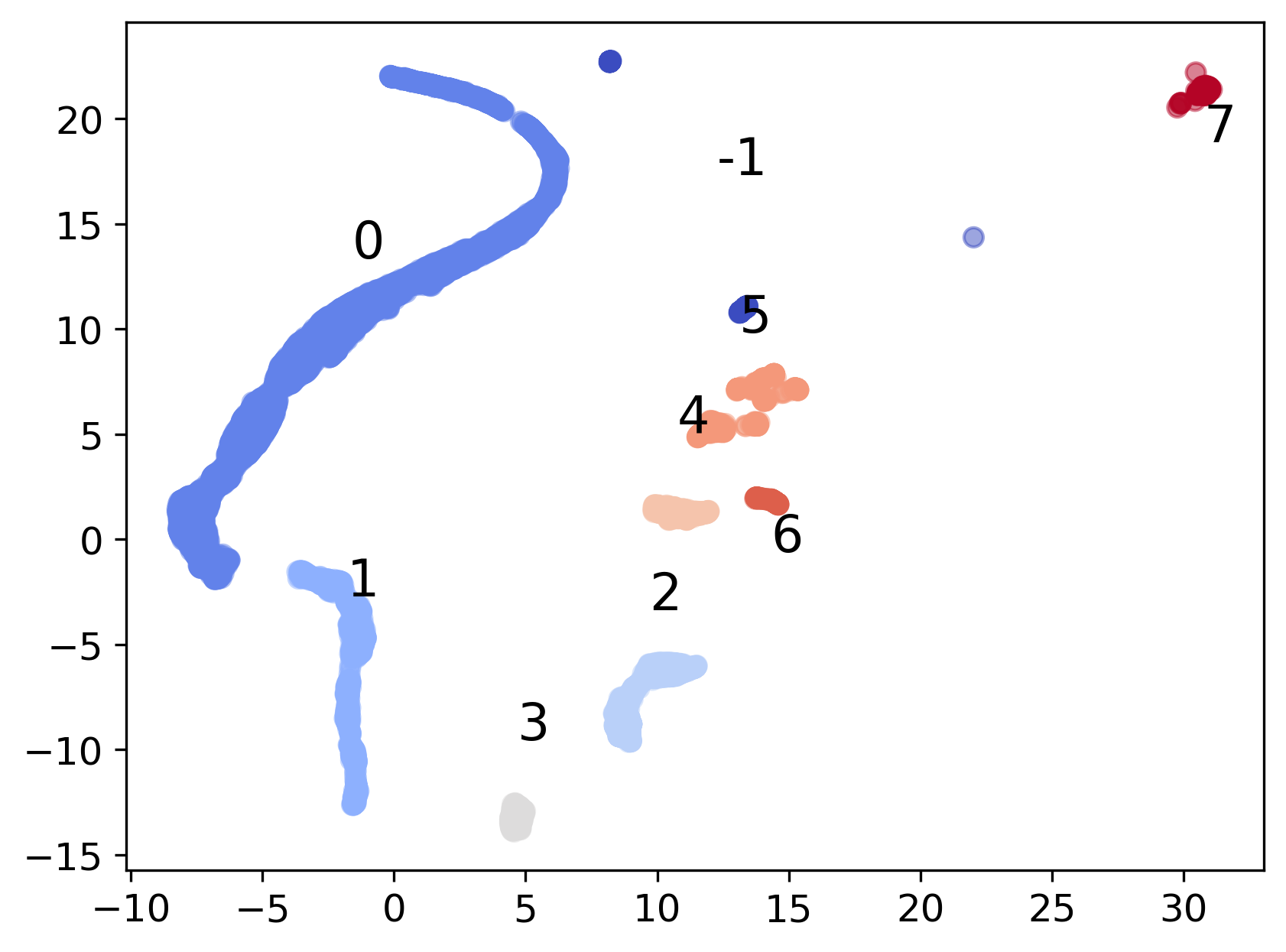}
        \caption{SCAL \\(Building 1) \\ $\tilde{\mathcal{U}}_{(-1)}$=\true \\SS=0.68721}
    \end{subfigure}
    \rulesep
    \begin{subfigure}[b]{0.23\linewidth}
        \centering
        \includegraphics[width=\textwidth]{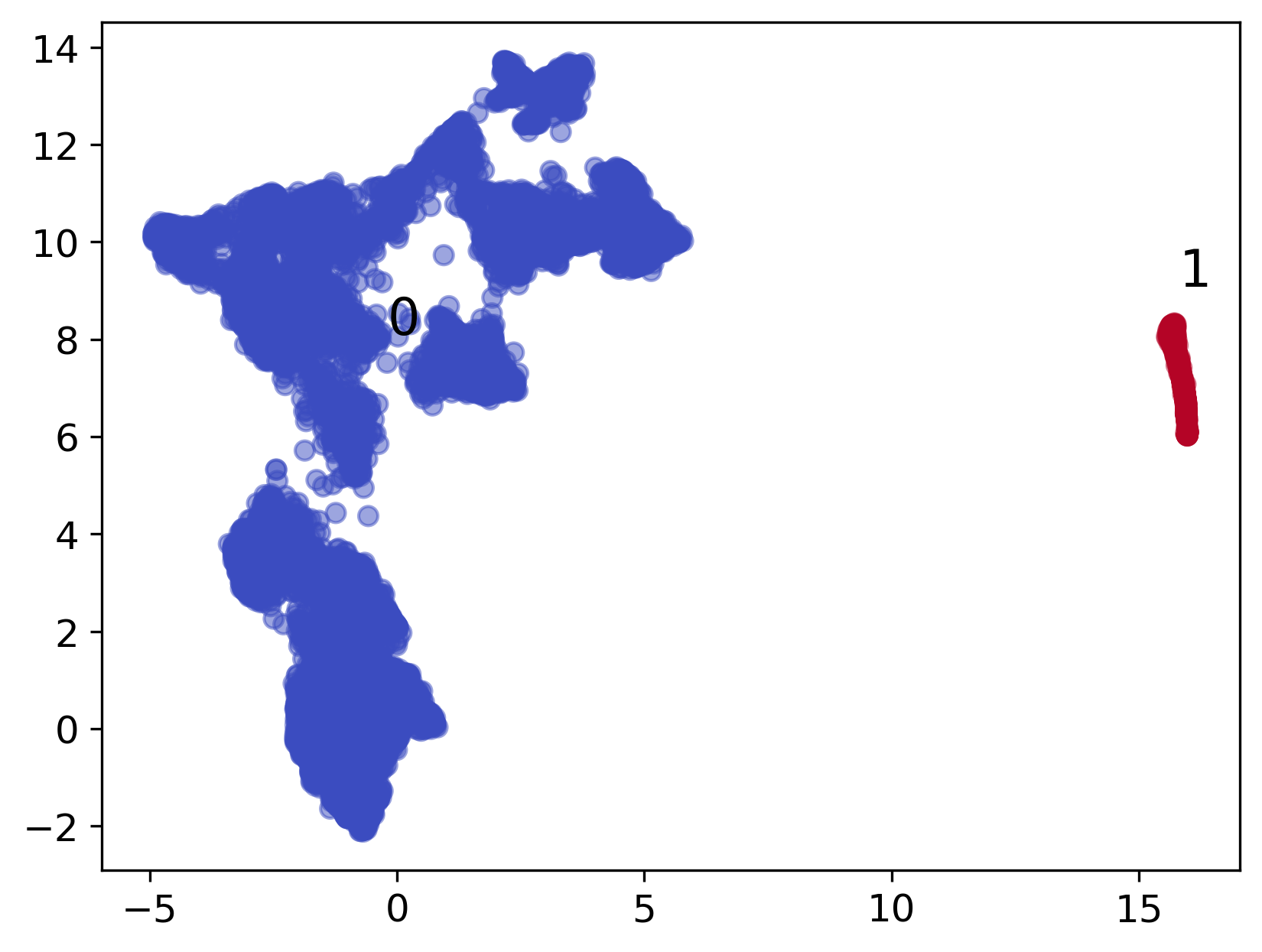}
        \caption{AHT \\(Building 2) \\ $\tilde{\mathcal{U}}_{(-1)}$=\false \\SS=0.71530}
    \end{subfigure}
    \begin{subfigure}[b]{0.24\linewidth}
        \centering
        \includegraphics[width=\textwidth]{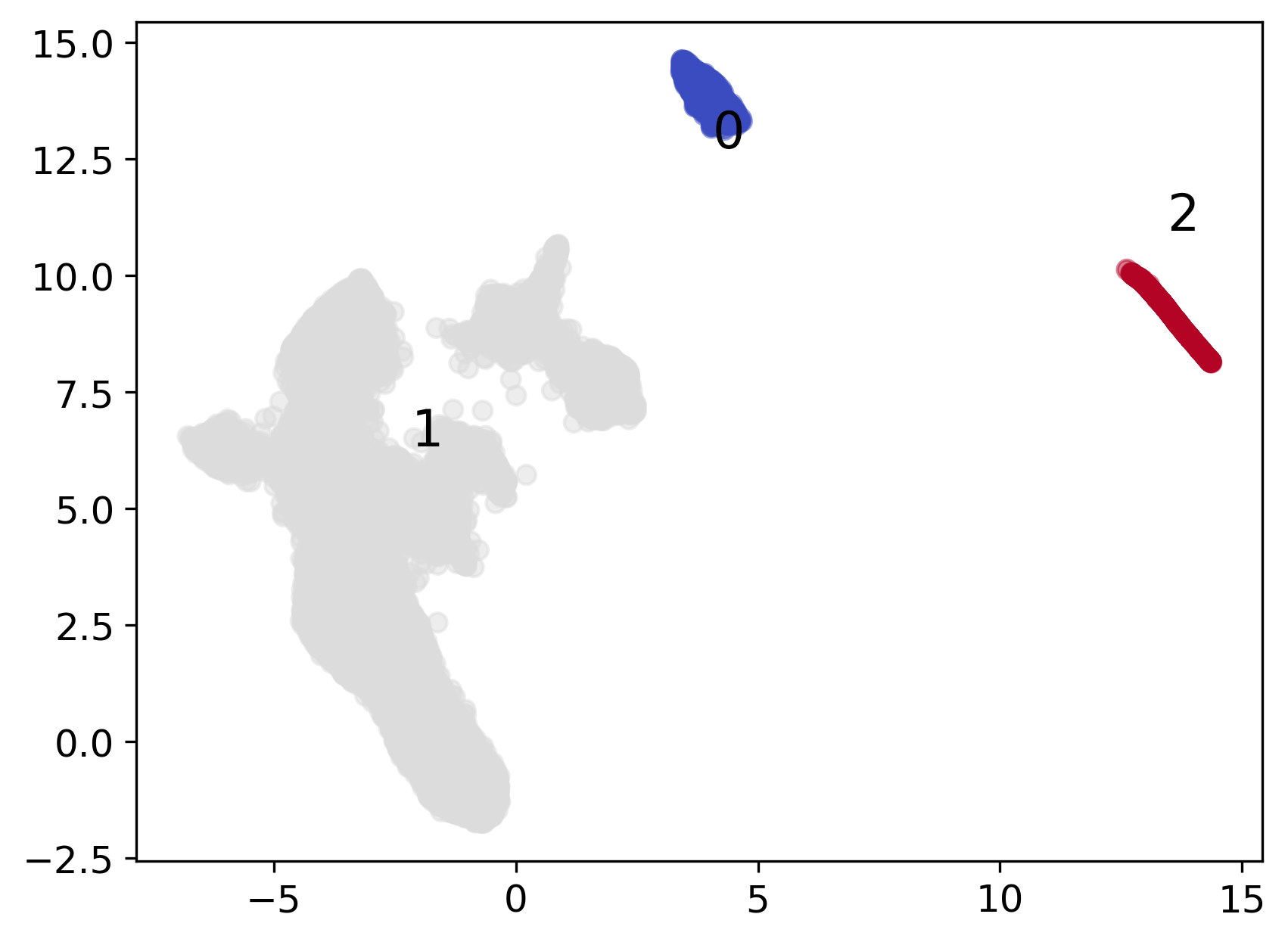}
        \caption{SCAL \\(Building 2) \\ $\tilde{\mathcal{U}}_{(-1)}$=\false \\SS=0.72695}
    \end{subfigure}
    \par\medskip
        \begin{subfigure}[b]{0.24\linewidth}
        \centering
        \includegraphics[width=\textwidth]{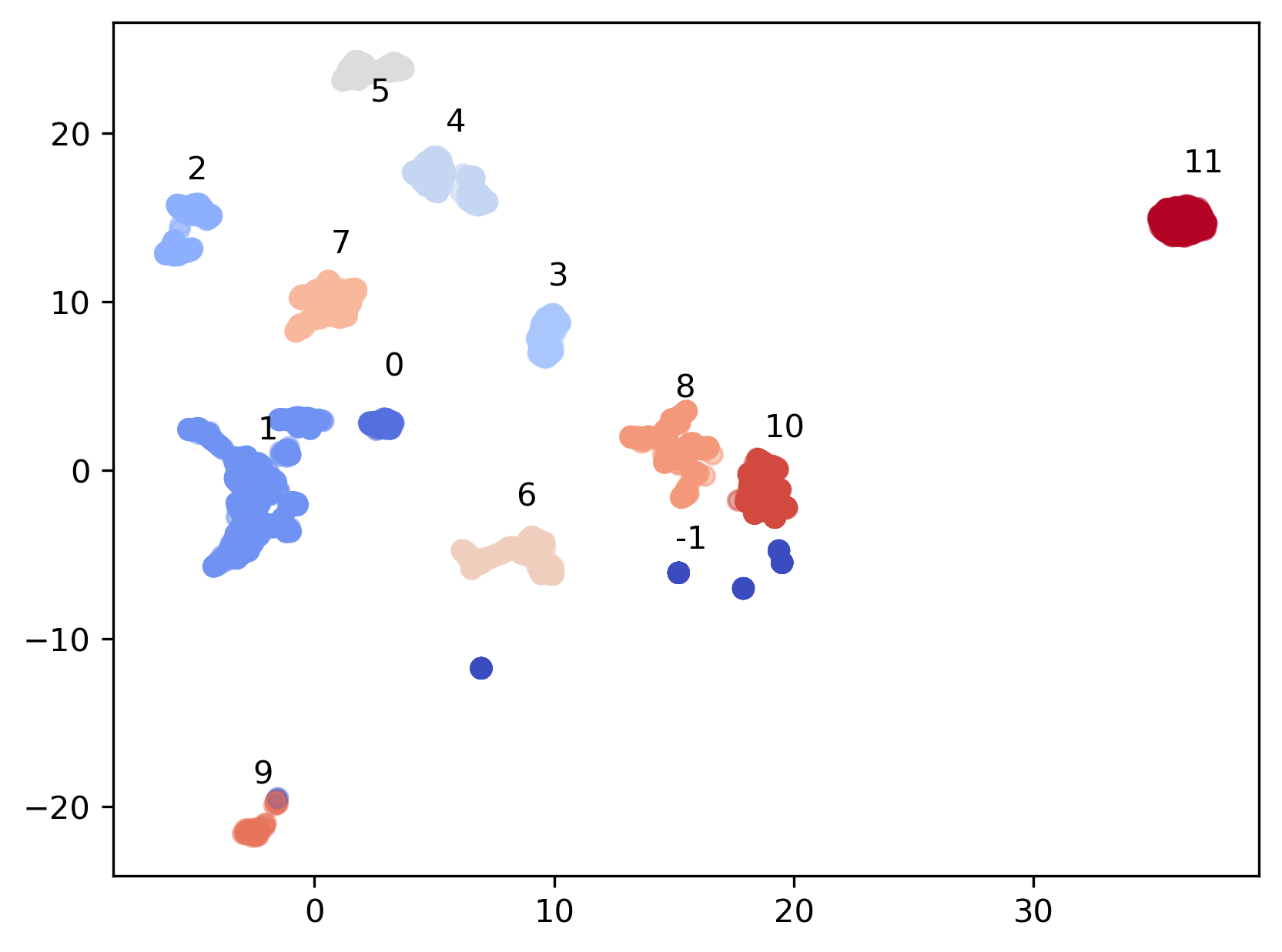}
        \caption{AHT \\(Building 3) \\ $\tilde{\mathcal{U}}_{(-1)}$=\true \\SS=0.77496}
    \end{subfigure}
    \begin{subfigure}[b]{0.24\linewidth}
        \centering
        \includegraphics[width=\textwidth]{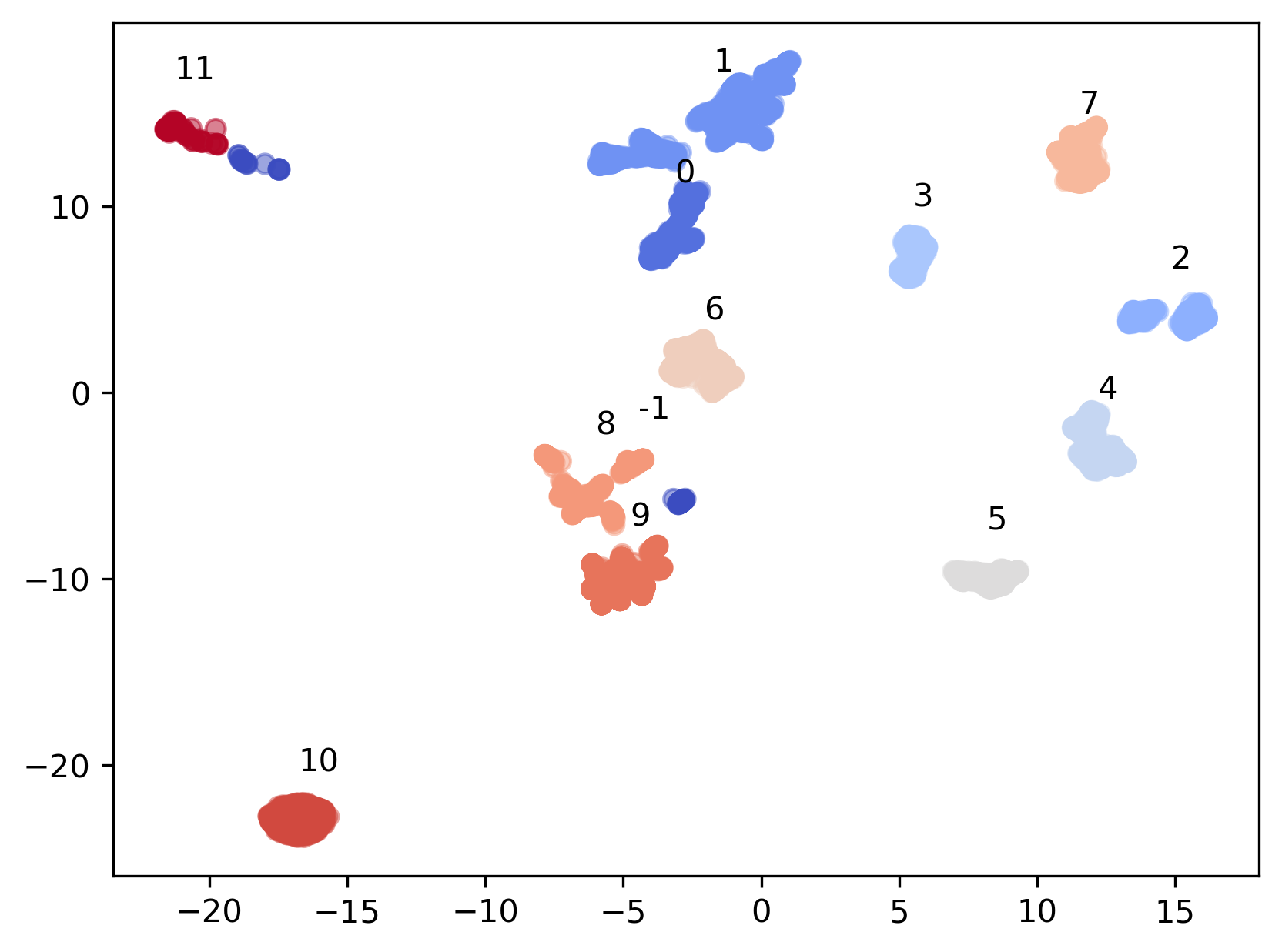}
        \caption{SCAL \\(Building 3) \\ $\tilde{\mathcal{U}}_{(-1)}$=\true \\SS=0.86414}
    \end{subfigure}
    \rulesep
    \begin{subfigure}[b]{0.24\linewidth}
        \centering
        \includegraphics[width=\textwidth]{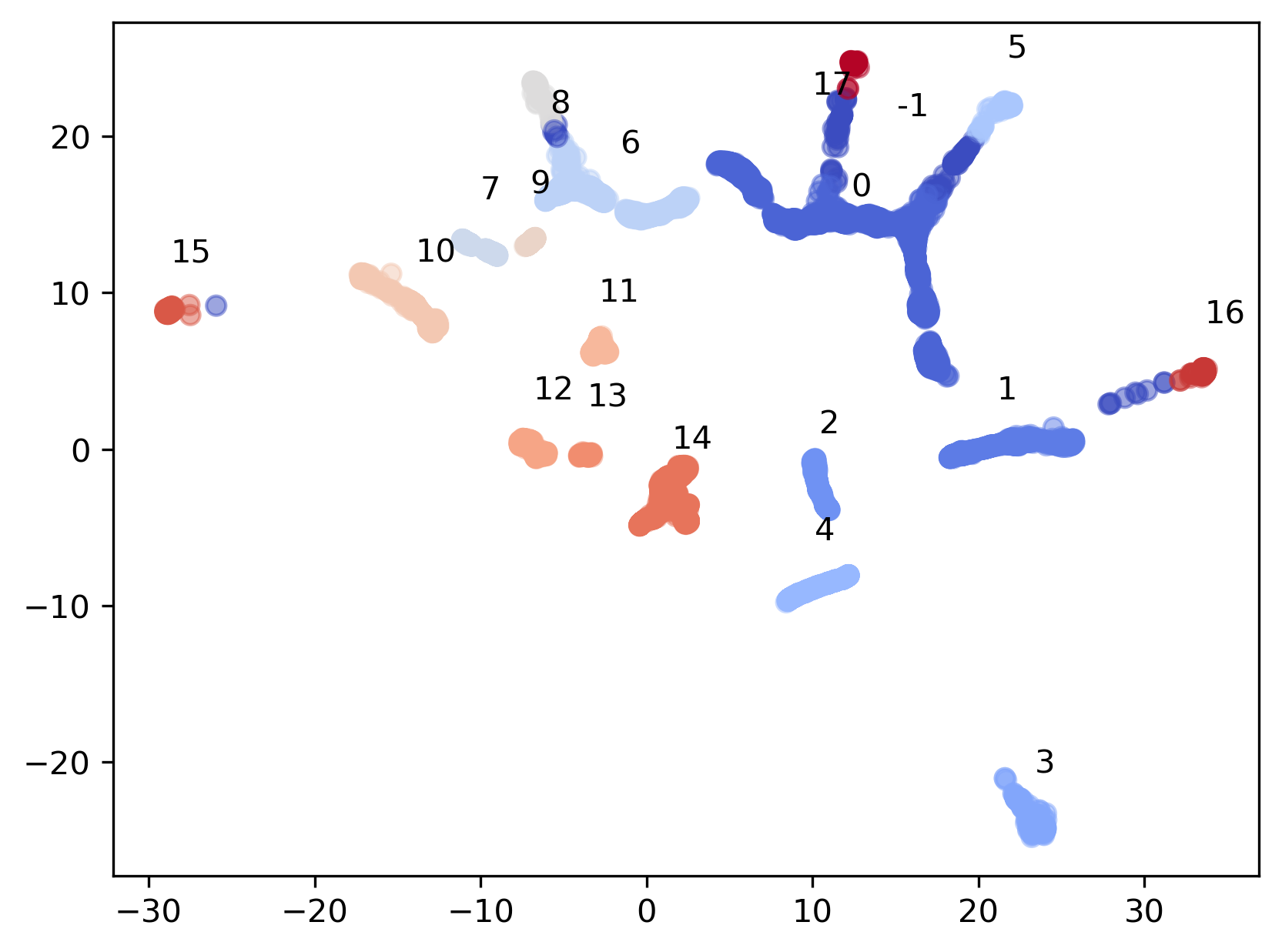}
        \caption{AHT \\(Building 4) \\ $\tilde{\mathcal{U}}_{(-1)}$=\true \\SS=0.21900}
    \end{subfigure}
    \begin{subfigure}[b]{0.24\linewidth}
        \centering
        \includegraphics[width=\textwidth]{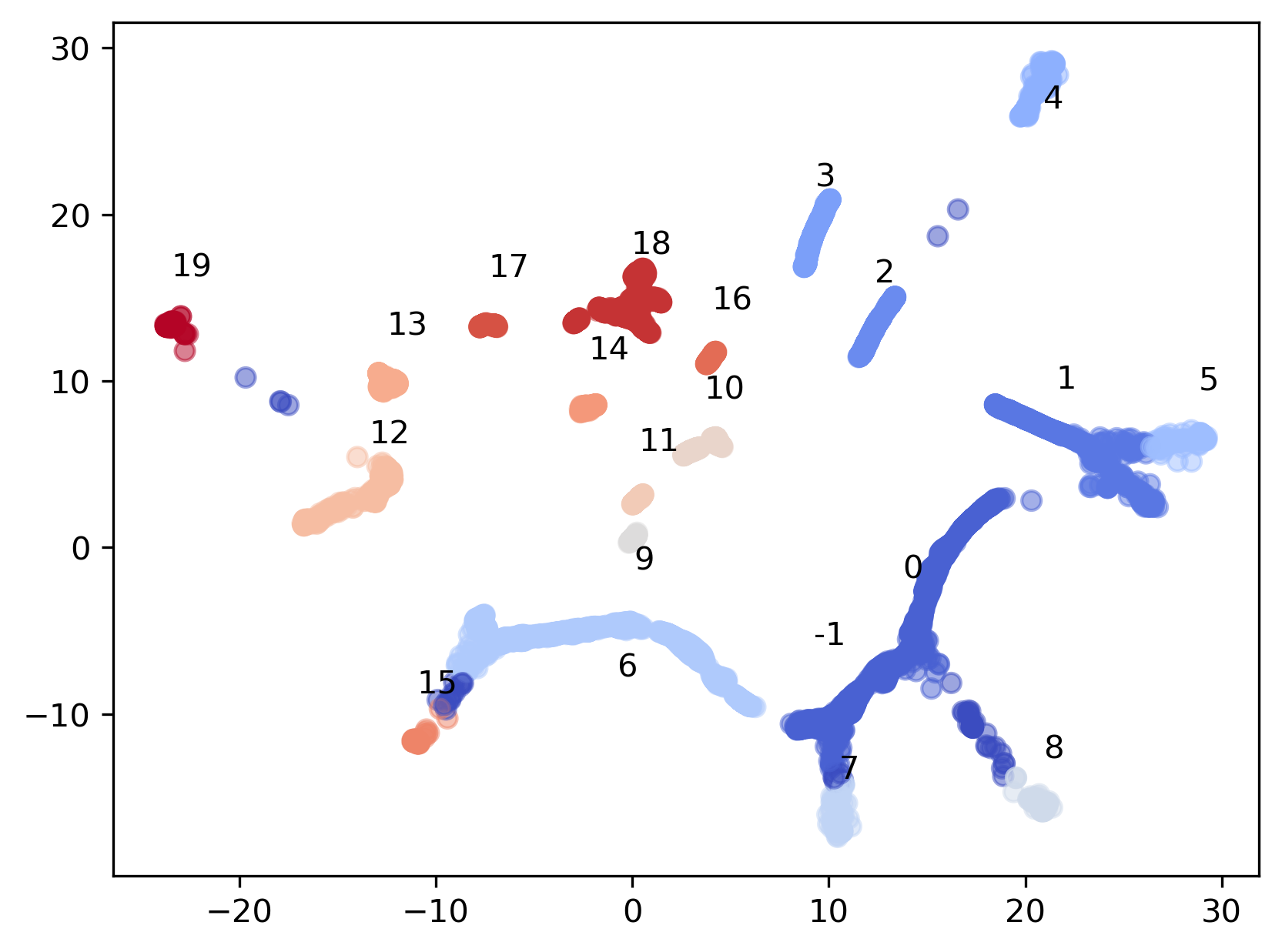}
        \caption{SCAL \\(Building 4) \\ $\tilde{\mathcal{U}}_{(-1)}$=\true \\SS=0.25013}
    \end{subfigure}
    
    \caption{SHAP clusters embedding of AHT and SCAL on the four different buildings' training set. Each cluster goes with its index, where -1 indicates the noise cluster. SCAL improves the cluster's quality with a higher silhouette score (SS) and the presence of the noise cluster, where the noise cluster presents ($\true$) and does not present (\false).}
    \label{fig:building_cluster_2}
    \vspace{-30pt}
\end{figure}

\subsubsection{Data error detection case}
\begin{SCfigure}[][t!]
    \centering
    \includegraphics[width=.35\linewidth]{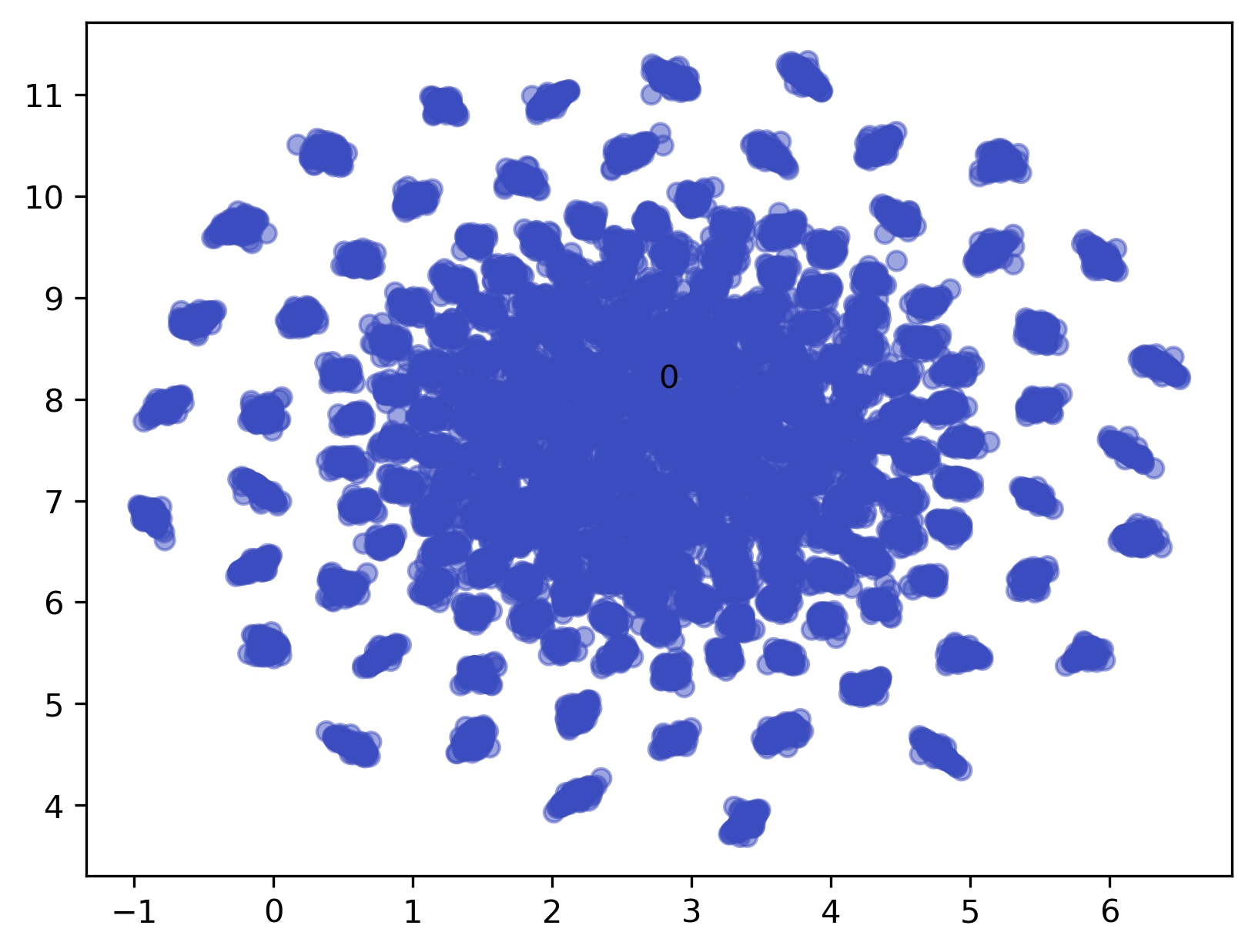}
    \vspace{-20pt}
    \caption{SHAP clusters of building 12 on the training set. In this case, SCAL can detect the data error where only one cluster appears.}
    \label{fig:detect_error}
    \vspace{15pt}
\end{SCfigure}

Like its category counterparts (Figure~\ref{fig:cluster_building}), SHAP clustering for building 12 displayed a single cluster in the explanation space (Figure~\ref{fig:detect_error}), indicating the model's reliance on a unique rule to classify all data points and its struggle to interpret the training dataset. Changing model parameters didn't affect this clustering, further emphasizing the model's difficulty in obtaining valuable insights. Detailed analysis of the training set unveiled numerous errors, including missing or incorrect entries, likely due to the data's real-world origin from a smart meter device. Thus, SHAP clustering offers crucial insights for diagnosing dataset issues, guiding users to enhance dataset features and fix problems, underscoring the efficacy of our approach in improving model performance.

\section{Transferability to other Use Cases}
In this section, we extend the application of the SCAL method beyond its original context, by examining its performance on two additional use cases. The first is a classification problem using the Financial Distress dataset, while the second is a regression problem using the Power dataset. These experiments serve to verify SCAL's adaptability and efficacy across varied datasets and problem types.

\subsection{Financial Distress data set (Classification problem)}
We experiment to test the SCAL method's performance on a multivariate time series classification problem using the Financial Distress data set~\cite{Financia52:online}, which aims to predict financial distress for a sample of companies and is considered imbalanced, with 136 financially distressed companies and 286 healthy ones.
The results of applying SCAL to the XGBoost model are shown in Table~\ref{tab:quantitative_results_fin}. Although our SCAL method was initially designed for regression problems, we explore its effectiveness in a classification context. In this experiment, the tuning XGBoost hyperparameters algorithm produces a model with \texttt{max\_depth}=9 and \texttt{gamma}=0, while the SCAL adapts \texttt{max\_depth}=8 and \texttt{gamma}=0.001. As a classification problem, Table~\ref{tab:quantitative_results_fin} displays the performance of the AHT and SCAL under accuracy.
By observing the silhouette score of the SHAP clusters in the training set's embedding, the SCAL method suggests using a shallower tree, which results in improved performance on the test set. 
\vspace{-10pt}
\begin{table}[htb!]
\setlength{\tabcolsep}{5pt}
\centering\small
\caption{The quantitative comparison between the AHT and SCAL on the Financial Distress data set. The best results are in bold. We only consider the silhouette score (SS) in the training set. The arrow indicates that the lower/higher, the better. Noise cluster presents (True) and does not present (False).}
\resizebox{.7\linewidth}{!}{%
\begin{tabular}{ccccccccc}
\toprule
\multirow{2}{*}[-2pt]{Data set} & \multicolumn{3}{c}{AHT} & \multicolumn{3}{c}{SCAL}  \\
\cmidrule(lr){2-4} \cmidrule(lr){5-7}
 & $\tilde{\mathcal{U}}_{(-1)}$ & $SS \uparrow$  & $Acc\% \uparrow$ & $\tilde{\mathcal{U}}_{(-1)}$ & $SS \uparrow$  & $Acc\% \uparrow$ \\
\midrule
Training  & $\false$ & 0.70096 & 100  & $\false$ & \textbf{0.72854}    &   100  \\
\midrule
Test & - & - & 95.918 &   -& -  &  \textbf{96.190} \\
\bottomrule
\end{tabular}
}
\label{tab:quantitative_results_fin}
\vspace{-10pt}
\end{table}

\subsection{Power data set (Regression problem)}
We evaluate SCAL on a public regression problem, namely the Power data set. This data set measures electric power consumption in one household with a one-minute sampling rate over a period of nearly four years. It shares several similarities in the domain and features with our recorded data set, such as the distribution difference between training and test set and missing values.
The results are shown in Table~\ref{tab:quantitative_results_power}.
In this experiment, the hyperparameters with AHT are \texttt{max\_depth}=8 and \texttt{gamma}=0, while the SCAL adapts \texttt{max\_depth}=7 and \texttt{gamma}=0.01.
The silhouette score of the SHAP clusters in the training set indicates that SCAL produces better-quality clusters. 
In the test set, SCAL outperforms AHT in both RMSE and $r^2$ metrics, demonstrating its effectiveness on this regression problem.

%\begin{SCfigure}[][h!]
    %\subfloat[\centering AHT \\($\tilde{\mathcal{U}}_{(-1)}$=\true, SS=0.38326)]{
        %\includegraphics[width=0.32\hsize]{images/power_cluster_train_HTM.png}\label{fig:aht_power}
    %}
    %\hfill
    %\subfloat[\centering SCAL \\($\tilde{\mathcal{U}}_{(-1)}$=\false, SS=0.41241)]{\includegraphics[width=0.32\hsize]{images/power_cluster_train_scal.png}\label{fig:scal_power}}
    %\caption{SHAP clusters embedding of the AHT and SCAL on the Power training set. SCAL improves the quality of the clusters by having a higher silhouette score (SS).}
    %\label{fig:power_cluster}
%\end{SCfigure}
\vspace{-20pt}
\begin{table}[h!]
\setlength{\tabcolsep}{5pt}
\centering\small
\caption{The quantitative comparison between the AHT and SCAL on the Power data set. The best results are in bold. We only consider the silhouette score (SS) in the training set. The arrow indicates that the lower/higher, the better. Noise cluster presents ($\true$) and does not present ($\false$).}
\raisebox{-0.5\height}{%
\resizebox{\linewidth}{!}{%
\begin{tabular}{ccccccccc}
\toprule
\multirow{2}{*}[-2pt]{Data set} & \multicolumn{4}{c}{AHT} & \multicolumn{4}{c}{SCAL}  \\
\cmidrule(lr){2-5} \cmidrule(lr){6-9}
 & $\tilde{\mathcal{U}}_{(-1)}$ & $SS \uparrow$  & $RMSE \downarrow$ & $r^2 \uparrow$ & $\tilde{\mathcal{U}}_{(-1)}$ & $SS \uparrow$  & $RMSE \downarrow$ & $r^2 \uparrow$ \\
\midrule
Training  & $\true$ & 0.38326 & 0.48246 &  0.82864  & $\true$ & \textbf{0.41241} &  0.53600 &  0.78851  \\
\midrule
Test & - & - & 0.75378 & 0.64520 &  -& -  &  \textbf{0.70856} &  \textbf{0.68649} \\
\bottomrule
\end{tabular}
}
}
\label{tab:quantitative_results_power}
\end{table}
\vspace{-15pt}

\section{Conclusion and Future Work}
Our paper introduces a method to enhance energy consumption prediction models by adapting to data distribution shifts. It integrates XAI techniques into an adaptive learning framework, using SHAP clustering to create an interpretable explanation space for model refinement. Our approach not only shows flexibility to data shifts but also potential applicability across various domains, as demonstrated in our experiments. Despite increased computational complexity, outlined in Appendix \ref{appendix:computational_complexity}, the robustness and interpretability benefits of our method offer an appealing trade-off.
Looking ahead, our research may explore additional XAI techniques and advanced clustering algorithms, and consider adapting our method to different machine learning models. We aim to develop a formula incorporating our three key quality metrics for further optimization and investigate a more self-adaptive learning process. Future work will also be informed by insights from our ablation study (Appendix \ref{appendix:ablation_study}).
In conclusion, we provide a promising framework for integrating XAI into adaptive learning, potentially aiding future research in fields requiring precise, interpretable predictions.

\bibliographystyle{splncs04}
\bibliography{main.bib}

\appendix
\section{Appendix}

\subsection{Computational Complexity}
\label{appendix:computational_complexity}
This section shows the computational complexity of the SCAL method. We use a computational resource with the A5000 NVIDIA GPU and 32GB RAM. In the aspect of time complexity, we show the computational time by each step in the process, as shown in Table~\ref{tab:computational_time}. We divide the computational time of SCAL into the initialization step and the adaptation step because the tuning XGBoost hyperparameters algorithm is only used as the model initializer, which is usually the longest step in the whole process. The time complexity depends on the size of the data set, where each building in our Energy consumption data set contains nearly 30000 samples. The initialization step usually takes 200 seconds, and each adaption step takes nearly 68 seconds. Also, the total training time of SCAL depends on how many adaption steps are taken to converge to the final SCAL. In our experiments, the model converges after X adaption steps on average.
It is important to consider the trade-off between computational time and model performance. Although the SCAL method may require more time for initialization and adaptation compared to the traditional AHT, it leads to better generalization and provides additional insights into the explanation space. This extra information can be invaluable for model refinement and adaptation, justifying the increased computational complexity.
\vspace{-5pt}
\begin{SCtable}
\centering\scriptsize
\caption{The average computational time in seconds of each step in the SCAL method on each building in the Energy consumption training set.}
\begin{tabular}{lc}
\toprule
\textbf{Algorithm} & \textbf{Time(s)}\\
\midrule
Tuning XGBoost hyperparameters & 132.6909 \\ 
Training the XGBoost model &  0.93 \\
SHAP values calculation & 22.053 \\
Dimension reduction with UMAP & 43.4812 \\
Clustering with DBSCAN & 0.9237 \\
Adapting hyperparameters & 0.0001 \\
\midrule
Running time in the initialization step & 200.07 \\
Running time in each adaption step & 67.388 \\
\bottomrule
\end{tabular}

\label{tab:computational_time}
\end{SCtable}

\subsection{Ablation study}
\label{appendix:ablation_study}
\begin{figure}[t!]
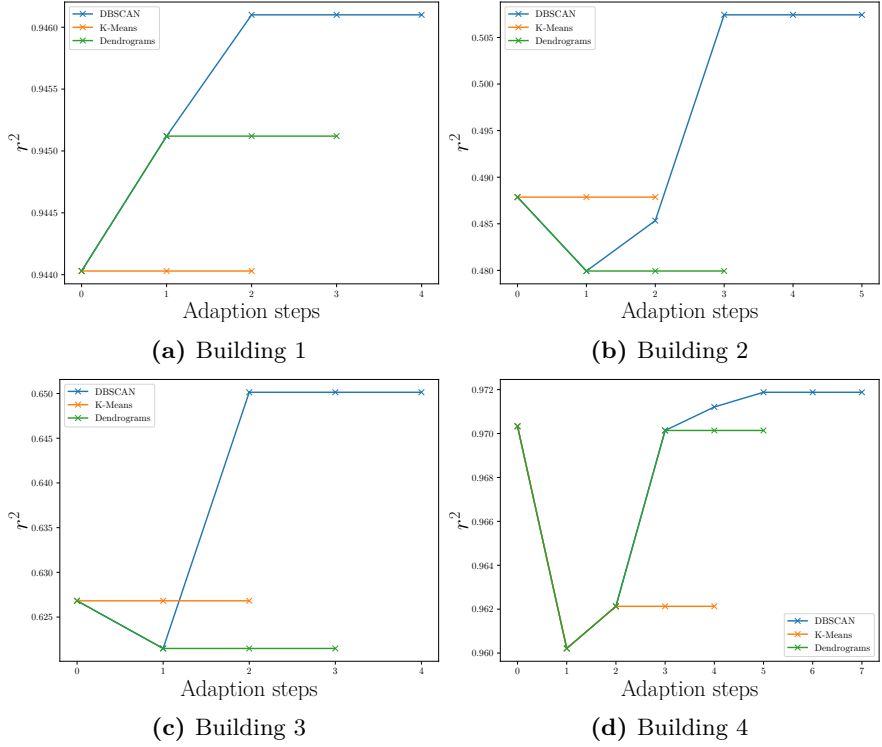

    %\vspace{-150pt}
    \captionsetup[subfigure]{justification=centering}
    \begin{subfigure}[b]{0.48\linewidth}
        \centering
        \scalebox{0.36}{\input{images/ablation_plots.pgf}}
        \caption{Building 1}
        \label{fig:ablation_study_1}
    \end{subfigure}%
    \begin{subfigure}[b]{0.48\linewidth}
        \centering
        \scalebox{0.36}{\input{images/ablation_plots_2.pgf}}
        \caption{Building 2}
        \label{fig:ablation_study_2}
    \end{subfigure}
    % \par\medskip
    \begin{subfigure}[b]{0.48\linewidth}
        \centering
        \scalebox{0.36}{\input{images/ablation_plots_3.pgf}}
        \caption{Building 3}
        \label{fig:ablation_study_3}
    \end{subfigure}%
    \begin{subfigure}[b]{0.48\linewidth}
        \centering
        \scalebox{0.36}{\input{images/ablation_plots_4.pgf}}
        \caption{Building 4}
        \label{fig:ablation_study_4}
    \end{subfigure}
    \caption{The goodness-of-fit of the SCAL on the test set under different clustering algorithms: DBSCAN (blue), $K$-Mean (orange), Dendrogram (green). The algorithm which achieves the higher $r^2$ in a shorter number of adaption steps is better.}
    \label{fig:ablation_study}
    
\end{figure}

In this section, we conduct an ablation study to investigate the impact of dimensionality reduction and clustering algorithms on the performance of the SCAL method. We compare three clustering methods: Partitional Clustering with $K$-means, Hierarchical Clustering with Dendrogram, and Density-Based Clustering with DBSCAN, on the test sets of four representative buildings. As shown in Fig.~\ref{fig:ablation_study}, DBSCAN outperforms $K$-means and Dendrogram clustering algorithms, achieving the highest $r^2$ on the test sets with the fewest adaptation steps.
$K$-means clustering is not designed to identify noise clusters, as it aims to partition the dataset into $k$ predefined, distinct, non-overlapping clusters. Furthermore, outliers can influence centroids, potentially resulting in their formation of separate clusters rather than being disregarded~\cite{im2020fast}. Dendrogram clustering can identify noise points as single-point clusters or clusters with very few points when cut at a certain level, enabling it to capture noise clusters~\cite{elkin2020mergegram}. However, as demonstrated in Fig.~\ref{fig:ablation_study_3}, Dendrogram remains sensitive to noise and outliers, which adversely affects noise cluster identification consistency and cluster quality as measured by the silhouette score. This may lead to suboptimal performance on the test set.
DBSCAN is capable of identifying noise clusters due to its noise resistance and ability to handle clusters of varying shapes and sizes—two critical features for accurately estimating the silhouette score and detecting the presence of noise clusters, which ultimately enhance the overall performance of the SCAL method. Consequently, we select DBSCAN as our default clustering algorithm.

\end{document}